\documentclass{llncs}
\usepackage{makeidx}  
\usepackage{amsmath, mleftright}    
\usepackage{graphicx}   
\usepackage[lofdepth,lotdepth]{subfig}
\usepackage{tabularx}
\usepackage[font=footnotesize,labelfont=footnotesize]{caption}
\usepackage{verbatim}   
\usepackage{color}      
\usepackage{hyperref} 
\usepackage{multirow}
\usepackage{multicol}
\usepackage{graphicx}
\usepackage{rotating}
\usepackage{array}
\usepackage{xspace}
\usepackage{indentfirst}
\usepackage{amssymb}
\usepackage{float}
\usepackage{algorithm}
\usepackage[algo2e]{algorithm2e}
\usepackage{algorithmic}
\usepackage{comment}
\usepackage{sidecap}
\usepackage{booktabs}
\usepackage{breqn}
\usepackage{cite}
\usepackage{mathtools}
\usepackage{graphicx}
\graphicspath{{./figures/}}
\usepackage{caption}
\usepackage{float}
\usepackage{lipsum}
\usepackage{xcolor}
\usepackage{siunitx}
\usepackage{xspace}
\usepackage{bm}
\usepackage{gensymb}

\usepackage[inline]{enumitem}
\usepackage{paralist}
\usepackage{graphicx}

\usepackage{subfig}

\newcommand{\inv}{^{\raisebox{.2ex}{$\scriptscriptstyle-1$}}}
\begin{document}
\frontmatter          
\pagestyle{headings}  
\addtocmark{Hamiltonian Mechanics} 
\mainmatter              
\title{Extrinsic Calibration of LiDAR, IMU and Camera}
\titlerunning{Extrinsic Calibration of LiDAR, IMU and Camera}  
%
%

\author{Subodh Mishra \and Srikanth Saripalli}

\authorrunning{Mishra et al.} 
%
\tocauthor{Subodh Mishra, Srikanth Saripalli}
\institute{Texas A$\&$M Univeristy, College Station, USA\\
\email{subodh514@tamu.edu}}

\maketitle              
\begin{abstract}
In this work we present a novel method to jointly calibrate a sensor suite consisting a 3D-LiDAR, Inertial Measurement Unit (IMU) and Camera under an Extended Kalman Filter (EKF) framework. We exploit pairwise constraints between the 3 sensor pairs to perform EKF update and experimentally demonstrate the superior performance obtained with joint calibration as against individual sensor pair calibration.\\
\vspace{-0.75cm}
\end{abstract}
\vspace{-1.25cm}
\begin{figure}[h]
    \centering
    \includegraphics[scale=0.25]{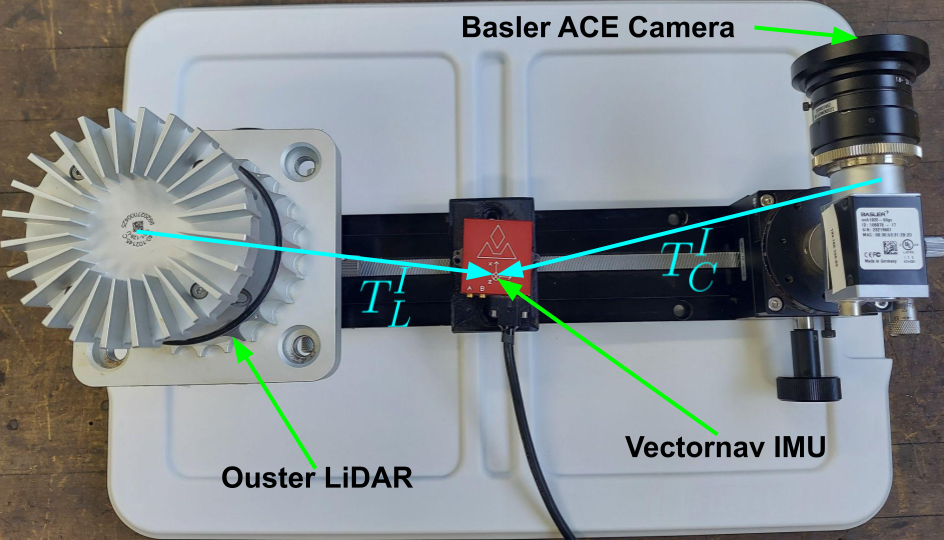}
    \caption{\scriptsize{\textbf{Lidar IMU Camera System:} The goal of this work is to determine LiDAR-IMU extrinsics $T^{I}_{L}\in SE(3)$ and Camera-IMU extrinsics $T^{I}_{C}\in SE(3)$. We perform our experiments with data from a sensor suite (Figure \ref{fig: lidarcameraimusetup}) consisting an Ouster 128 Channel LiDAR running at 10 Hz, a Basler Ace Camera $[1600 \times 1200]$ running at 30 Hz, and a Vectornav VN-300 IMU running at 400 Hz. We collect calibration data from these sensor while exciting the sensor suite along all directions and about all axes.}}
    \label{fig: lidarcameraimusetup}
\end{figure}
\vspace{-1.5cm}
\section{Introduction $\&$ Related Work}
Autonomous Systems use LiDARs, IMUs and Cameras together to overcome an individual sensor's shortcoming by leveraging the advantages of another. Extrinsic calibration of sensors is performed so that measurements can be represented in the same spatial frame of reference for sensor fusion.  Although there are several contributions for performing LiDAR-camera (\cite{subodhIROS2020}), camera-IMU (\cite{Stergios},  \cite{kalibr}), IMU-LiDAR (\cite{lincalib1}, \cite{SubodhMFI2021}) extrinsic calibration, joint calibration of LiDAR-IMU-Camera is rarely addressed. Deriving motivation from previous works like \cite{subodhIROS2020}, \cite{Stergios}, \cite{SubodhMFI2021}, we present our contribution on joint extrinsic calibration of LiDAR-IMU-Camera system (Figure \ref{fig: lidarcameraimusetup}).

\section{Problem Formulation}
\vspace{-0.5cm}
Our goal is to determine the spatial 6 DoF separation, \emph{viz.} the extrinsic calibration $T^{I}_{L} \& T^{I}_{C} \in SE(3)$ between an IMU and a LiDAR/Camera,  by simultaneously exploiting measurement constraints between all the sensor pairs under an EKF based estimation framework. We not only use the residuals previously used in \cite{Stergios} and \cite{SubodhMFI2021} to estimate $T^{I}_{L}$ and $T^{I}_{C}$ pairs respectively, but also exploit detection of a rectangular checkerboard calibration target in both LiDAR and camera to further constrain the optimization by introducing LiDAR-camera measurement residuals (\cite{subodhIROS2020}) into the EKF framework. In addition to the static extrinsic calibration parameters $T^{I}_{L}$ $\&$ $T^{I}_{C}$, the EKF also estimates the time-varying accelerometer $\&$ gyroscope biases($\hat{\textbf{b}}_{a, k}$, $\hat{\textbf{b}}_{g, k}$ $\in R^{3\times1}$), the pose ($T^{G}_{I_{k}} \in SE(3)$) $\&$ velocity($^{G}\hat{\textbf{v}}_{I_{k}} \in R^{3\times1}$) of the IMU. The EKF state propagation equations have been omitted here in interest of space and can be found in \cite{Stergios} $\&$ \cite{SubodhMFI2021}. The update steps are described briefly in Sections \ref{sec: cameraimuupdate}, \ref{sec: lidarimuupdate}, \ref{sec: cameralidarupdate}. 
\vspace{-0.5cm}
\subsection{Camera-IMU update}
\label{sec: cameraimuupdate}
\vspace{-0.25cm}
The details of Camera-IMU state update can be found in \cite{Stergios}. Briefly, the detected corners $z_{IC}$ of the calibration checkerboard are used as measurements to update Camera-IMU extrinsic calibration $T^{I}_{C}$ and other IMU states. The method involves calculation of measurement residuals $r_{IC} = z_{IC} - h(T^{G}_{I_k}, T^{I}_{C}, \mathbf{X}_{G})$ and the corresponding Jacobians $H^{z_{IC}}_{T^{G}_{I_k}, T^{I}_{C}}$ for state $\&$ covariance update. $h()$ is the camera projection model used to project the known 3D positions $\mathbf{X}_{G}$ of the checkerboard's corners using the best available estimates of $T^{G}_{I_k}, T^{I}_{C}$.
\vspace{-0.5cm}
\subsection{LiDAR-IMU update}
\label{sec: lidarimuupdate} 
\vspace{-0.25cm}
The details of LiDAR-IMU state update can be found in \cite{SubodhMFI2021}. Briefly, IMU measurements and the best available estimate of extrinsic calibration $T^{I}_{L}$ are used to remove motion distortion from the LiDAR scan before performing scan matching for LiDAR pose estimation. Ego motion distortion, which happens because of the sequential nature of measurement acquisition in spinning LiDARs, occurs during calibration data collection when the user motion excites the sensor suite along all directions, about all axes. The undistorted scans are used for scan matching based LiDAR pose estimation with the goal to perform motion based calibration \cite{Taylor201605} under an EKF paradigm. The measurement $z_{IL}$ in this case is the LiDAR pose resulting from scan matching, and the predicted measurement is $h(T^{G}_{I_k}, T^{I}_{L}) = T^{I}_{L} {\inv} T^{G}_{I_1} {\inv} T^{G}_{I_k} T^{I}_{L} \in SE(3)$, here $T^{G}_{I_1}$ is the global IMU pose corresponding to first LiDAR scan. The resulting measurement residuals $r_{IL} = z_{IL}  \ominus h(T^{G}_{I_k}, T^{I}_{L})$ and the corresponding Jacobians $H^{z_{IL}}_{T^{G}_{I_k}, T^{I}_{L}}$ are used for state and covariance update. Here $\ominus$ operator denotes difference between quantities lying on-manifold.
\vspace{-0.5cm}
\subsection{Camera-LiDAR update}
\label{sec: cameralidarupdate}
\vspace{-0.35cm}
Camera-LiDAR update under EKF framework for performing LiDAR-IMU-Camera joint calibration is the novel contribution of this work. The measurement residual in this case involves alignment of calibration target plane parameters $[n_L, d_L]_{i}$, $[n_C, d_C]_{j}$ detected in LiDAR and camera sensors, at times $i$ and $j$ respectively. The measurement residuals are as given:
\begin{align}
r_{CL} = 
\begin{bmatrix}
r_{R}\\
r_{p}
\end{bmatrix} =
\begin{bmatrix}
n_{C} - {R^{I}_{C}}^{\top} R^{Ij}_{G} {R^{Ii}_{G}}^{\top} R^{I}_{L} n_{L}\\
n^{\top}_{C} p^{C_j}_{L_i} + d_{C} - d_{L}
\end{bmatrix} 
\label{eqn: cameralidarimuresidual}
\end{align}
Where $p^{C_j}_{L_i} = {R^{I}_{C}}^{\top} R^{I_j}_G ({R^{I_i}_G}^{\top} p^{I}_L + p^{G}_{I_i}) - {R^{I}_{C}}^{\top} (p^{I}_C + R^{I_j}_G p^{G}_{I_j})$. The residual $r_{CL}$ and the corresponding Jacobians $H^{r_{CL}}_{T^{G}_{I_i}, T^{G}_{I_j}, T^{I}_{L}, T^{I}_{C}}$ which are functions of state variables $T^{G}_{I_i}$, $T^{G}_{I_j}$, $T^{I}_{L}$, $T^{I}_{C}$, are used for state and covariance update. As the residual (Equation \ref{eqn: cameralidarimuresidual}) is a function of both $T^{I}_{L}$ and $T^{I}_{C}$, this EKF update step jointly updates/estimates Camera-IMU and LiDAR-IMU extrinsic calibration parameters. While this residual has been used for pair-wise calibration of Camera-LiDAR systems\cite{subodhIROS2020}, we have implemented it here to perform joint calibration of Camera, IMU and LiDAR together. 
\vspace{-0.5cm}
\section{Experiments}
\vspace{-0.25cm}
\subsection{Comparision of estimated $T^{I}_{C}$ with Kalibr \cite{kalibr}}
\vspace{-0.75cm}
\begin{table}[h]
\centering
\resizebox{\columnwidth}{!}{
\begin{tabular}{|c | c | c | c | c | c | c|} 
 \hline
 \textbf{Method} & \textbf{R}${}^\circ$ & \textbf{P}${}^\circ$ & \textbf{Y}${}^\circ$ & \textbf{x} [m] & \textbf{y} [m] & \textbf{z} [m]\\ 
 \hline\hline
 \textbf{Kalibr \cite{kalibr}} & 89.0547   & 1.0927 &  90.8185 & 0.0727 & 0.1254 &   -0.0756 \\ \hline
 \textbf{Individual Calibration \cite{Stergios}} & 89.1348 & 1.0712 &  90.7966 & 0.0698  &  0.1235 &  -0.0799\\ \hline
 \textbf{Joint Calibration (JC)} &  89.2269  &  1.0738  & 90.7073 & 0.0696  &  0.1228  & -0.0806 \\ \hline
\end{tabular}}
\caption{\small{$T^{I}_{C}$ estimated using the open source camera-IMU calibration tool box Kalibr \cite{kalibr}, our own implementation of \cite{Stergios} and the proposed method of joint calibration.}}
\label{table: comparisionwithkalibr}
\end{table}
\vspace{-1cm}
In the absence of ground truth we compare the estimation of $T^{I}_{C}$ from the proposed joint calibration framework with  $T^{I}_{C}$ estimated using Kalibr\footnote{\url{https://github.com/ethz-asl/kalibr}} Camera-IMU calibration tool, and also with our own implementation of Camera-IMU calibration method\footnote{\url{https://github.com/unmannedlab/camera_imu_calibration}} presented in \cite{Stergios}. The results have been tabulated in Table \ref{table: comparisionwithkalibr}. The camera-IMU calibration $T^{I}_{C}$ estimation from the proposed joint calibration approach is very close to the individual calibration from \cite{kalibr} $\&$ \cite{Stergios}, as evident from the low standard devitation of 0.0862$^{\circ}$, 0.0117$^{\circ}$, 0.0589$^{\circ}$ along roll, pitch $\&$ yaw and 0.0017 m, 0.0013 m, 0.0027 m along $x$, $y$ $\&$ $z$ axes.
\vspace{-0.5cm}
\subsection{Projection of LiDAR points on Camera by daisy chaining estimated $T^{I}_{C}$ $\&$ $T^{I}_{L}$}
\label{sec: qualitativeevaluation}
\begin{figure}%
    \centering
    \subfloat[\scriptsize{\textbf{Individual Calibration:} Projected LiDAR points misaligned}]{{\includegraphics[width=0.45\textwidth]{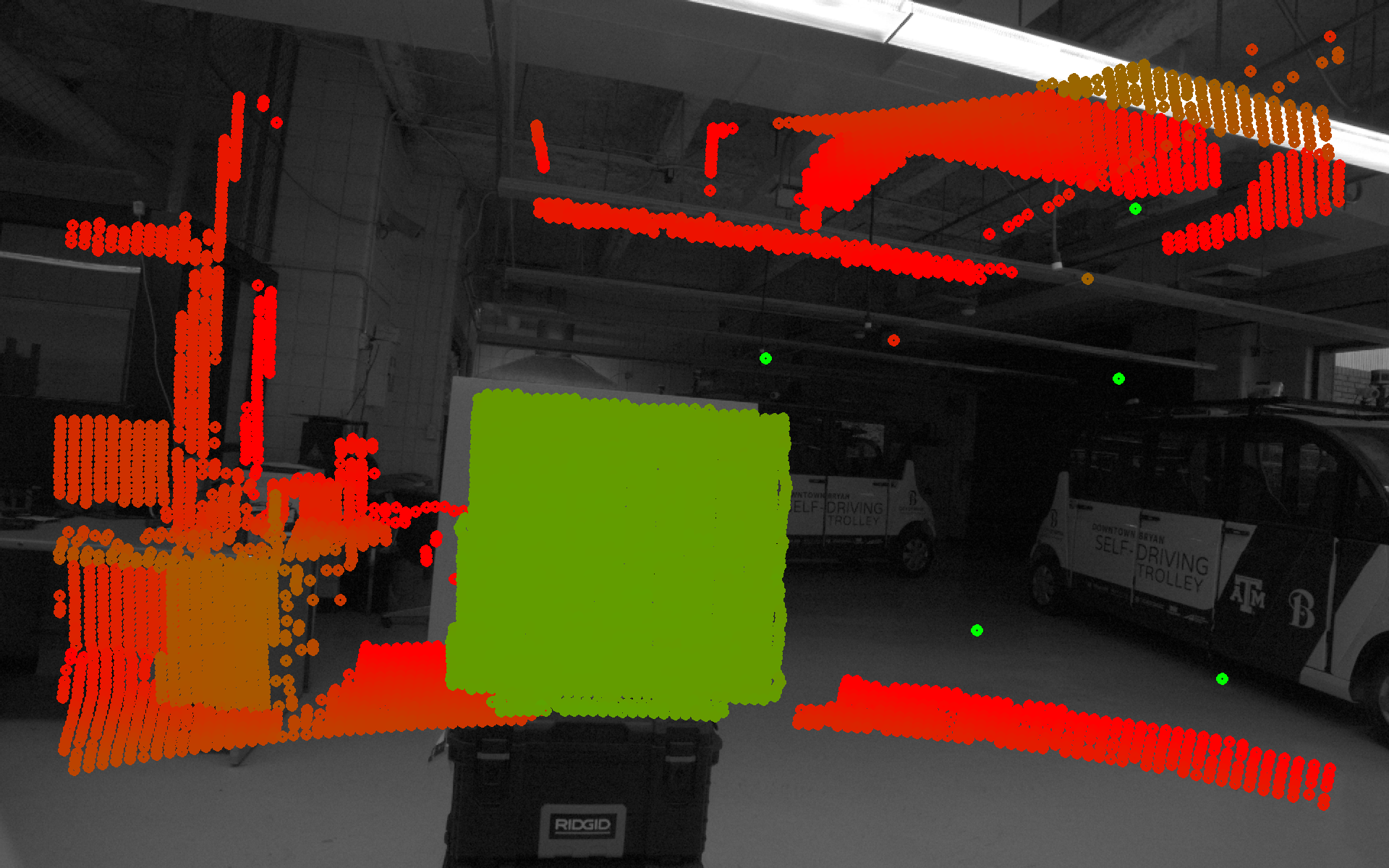} }\label{fig: daisychainedcalibration}}%
    \qquad
    \subfloat[\scriptsize{\textbf{Joint Calibration:} Projected LiDAR points well aligned}]{{\includegraphics[width=0.45\textwidth]{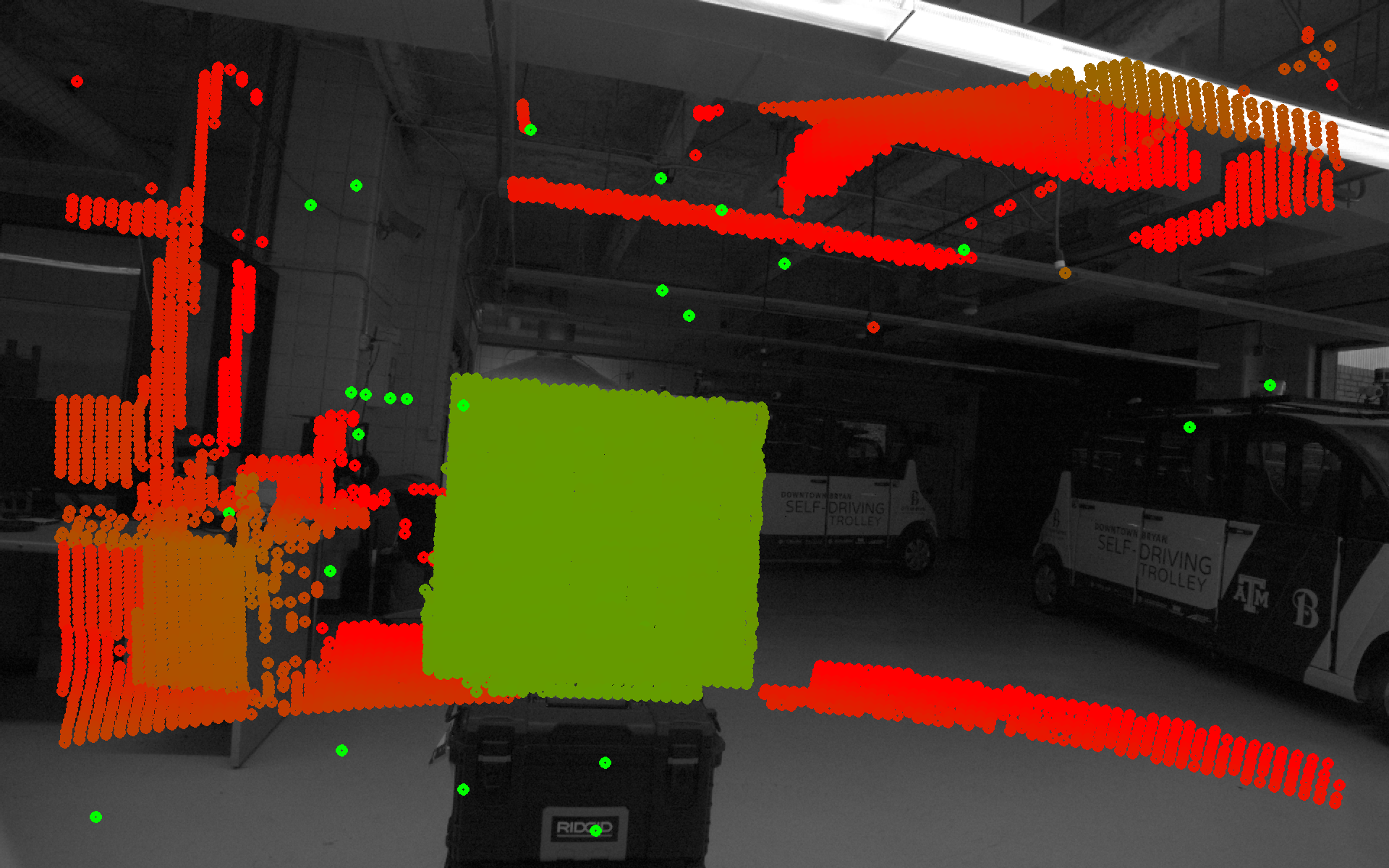} }\label{fig: jointcalibration}}%
    \caption{\scriptsize{\textbf{Qualitative Evaluation of Individual vs Joint Calibration}: } Daisy chaining estimated $T^{I}_{L}$ $\&$ $T^{I}_{C}$ for projecting LiDAR points on Camera Image.}%
    \label{fig: calibrationcomparison}%
\end{figure}

Although $T^{I}_{C}$ from various methods in Table \ref{table: comparisionwithkalibr} look very similar, the projection of LiDAR points ($\{X_L\}$) on image plane look different with individually (own implementation of \cite{Stergios} for camera-IMU, \cite{SubodhMFI2021} for LiDAR-IMU) and jointly estimated $T^{I}_{C}$ $\&$ $T^{I}_{L}$ \footnote{$T^{C}_{L} = T^{I}_{C} {\inv} T^{I}_{L}$} (using the proposed method). When $\{X_L\}$ are projected onto the image plane using individually estimated $T^{I}_{C}$ $\&$ $T^{I}_{L}$, the projected points do not align well with the corresponding image pixels, as evident in Figure \ref{fig: daisychainedcalibration} - the projected calibration board 3D points (green) do not align with the corresponding image pixels. Whereas, we do not see any such mis-alignment in Figure \ref{fig: jointcalibration} - where $\{X_L\}$ have been projected using jointly estimated $T^{I}_{C}$ $\&$ $T^{I}_{L}$.
\vspace{-1cm}
\vspace{-1.5cm}
\section{Conclusion}
\vspace{-0.3cm}
In this paper we presented a method to perform extrinsic calibration of a LiDAR-IMU-Camera system under the EKF framework. The extrinsic calibration $T^{I}_{C}$ for camera-IMU pair is similar across individual and joint calibration with low variance (Table \ref{table: comparisionwithkalibr}). However, the qualitative evaluation performed in Section \ref{sec: qualitativeevaluation} by projecting LiDAR points on image plane using jointly estimated $T^{I}_{C}$ $\&$ $T^{I}_{L}$ shows that the results from joint calibration are superior.
\vspace{-0.5cm}
\bibliographystyle{splncs04}
\bibliography{myRef}
%
%






\end{document}